# Ensemble CNN Networks for GBM Tumors Segmentation using Multi-parametric MRI

Ramy A. Zeineldin [1-3], Mohamed E. Karar[2], Franziska Mathis-Ullrich[3] and Oliver Burgert[1]

[1] Research Group Computer Assisted Medicine (CaMed), Reutlingen University, Germany
[2] Faculty of Electronic Engineering (FEE), Menoufia University, Egypt
[3] Health Robotics and Automation (HERA), Karlsruhe Institute of Technology, Germany
Ramy.Zeineldin@Reutlingen-University.DE

**Abstract.** Glioblastomas are the most aggressive fast-growing primary brain cancer which originate in the glial cells of the brain. Accurate identification of the malignant brain tumor and its sub-regions is still one of the most challenging problems in medical image segmentation. The Brain Tumor Segmentation Challenge (BraTS) has been a popular benchmark for automatic brain glioblastomas segmentation algorithms since its initiation. In this year, BraTS 2021 challenge provides the largest multi-parametric (mpMRI) dataset of 2,000 pre-operative patients. In this paper, we propose a new aggregation of two deep learning frameworks namely, DeepSeg and nnU-Net for automatic glioblastoma recognition in pre-operative mpMRI. Our ensemble method obtains Dice similarity scores of 92.00, 87.33, and 84.10 and Hausdorff Distances of 3.81, 8.91, and 16.02 for the enhancing tumor, tumor core, and whole tumor regions, respectively, on the BraTS 2021 validation set, ranking us among the top ten teams. These experimental findings provide evidence that it can be readily applied clinically and thereby aiding in the brain cancer prognosis, therapy planning, and therapy response monitoring. A docker image for reproducing our segmentation results is available online at (https://hub.docker.com/r/razeineldin/deepseg21).

## 1 Introduction

Glioblastomas (GBM), the most common and aggressive malignant primary tumors of the brain in adults, occur with ultimate heterogeneous sub-regions including the enhancing tumor (ET), peritumoral edematous/invaded tissue (ED), and the necrotic components of the core tumor (NCR) [1, 2]. Still, accurate GBM localization and its sub-regions in magnetic resonance imaging (MRI) are considered one of the most challenging segmentation problems in the medical field. Manual segmentation is the gold standard for neurosurgical planning, interventional image-guided surgery, follow-up procedures, and monitoring the tumor growth. However, identification of the GBM tumor and its sub-regions by hand is time-consuming, subjective, and highly dependent on the experience of clinicians.

The Medical Image Computing and Computer-Assisted Interventions Brain Tumor Segmentation Challenge (MICCAI BraTS) [3, 4] has been focusing on addressing this problem of finding the best automated tumor sub-region segmentation algorithm. The Radiological Society of North America (RSNA), the American Society of Neuroradiology (ASNR), and MICCAI jointly organize this year's BraTS challenge [2] celebrating its 10$^{th}$ anniversary. BraTS 2021 provides the largest annotated and publicly available multi-parametric (mpMRI) dataset [2, 5, 6] as a common benchmark for the development and training of automatic brain tumor segmentation methods.

Deep learning-based segmentation methods have gained popularity in the medical arena outperforming other traditional methods in brain cancer analysis [7-10], more specifically the convolutional neural network (CNN) [11] and the encoder-decoder architecture with skip connections, which are first introduced by the U-Net [12, 13]. In the context of the BraTS challenge, the recent winning contributions of 2018 [14], 2019 [15], and 2020 [16] extend the encoder-decoder pattern by adding variational autoencoder (VAE) in [14], two-stage cascaded U-Net [15], or using the baseline U-Net architecture with making significant architecture changes [16].

In this paper, we propose a fully automated CNN method for GDM segmentation based on an ensemble of two encoder-decoder methods, namely, DeepSeg [10], our recent deep learning framework for automatic brain tumor segmentation using two-dimensional T2 Fluid Attenuated Inversion Recovery (T2-FLAIR) scans, and nnU-Net [16], a self-configuring method for automatic biomedical segmentation. The remainder of the paper is organized as follows: Section 2 describes the BraTS 2021 dataset and the architecture of our ensemble method. Experimental results are presented in Section 3. This research work is concluded in Section 4.

## 2   Materials and Methods

### 2.1   Data

The BraTS 2021 training database [2] includes 1251 mpMRI images acquired from multiple institutions using different MRI scanners and protocols. For each patient, there are four mpMRI volumes: pre-contrast T1-weighted (T1), post-contrast T1-weighted (T1Gd), T2-weighted (T2), and T2-FLAIR, as shown in Fig. 1. Ground truth labels are provided for the training dataset only indicating background (label 0), necrotic and non-enhancing tumor core (NCR/NET) (label 1), peritumoral edema (ED) (label 2), and enhancing tumor (ET) (label 4). These labels are combined to generate the final evaluation of three regions: the tumor core (TC) of labels 1 and 4, enhancing tumor (ET) of label 4, and the whole tumor (WT) of all labels. Also, the BraTS 2021 includes 219 validation cases without any associated ground truth labels.

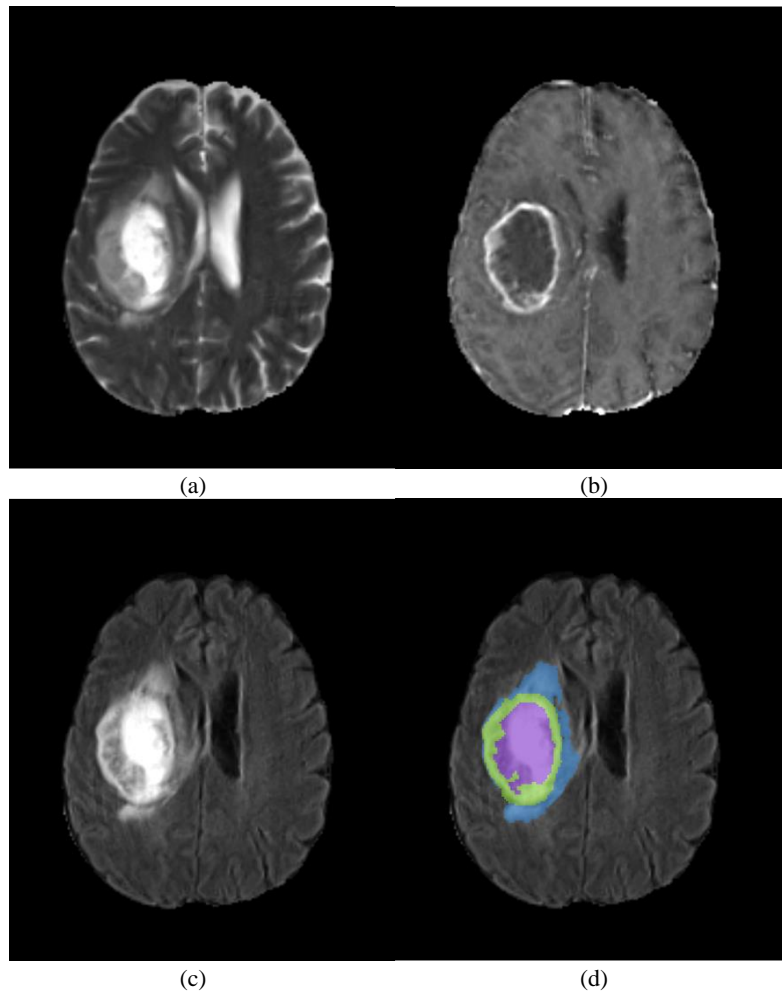

**Fig. 1.** A sample of the mpMRI BraTS 2021 training set. Shown are images slices in two different MRI modalities T2 (a), T1Gd (b), T2-FLAIR (c), and the ground truth segmentation (d). The color labels indicate Edema (*blue*), enhancing solid tumor (*green*), and non-enhancing tumor core, and necrotic core (*magenta*). Images were obtained by using the 3D Slicer software [17].

### 2.2 Data Pre-processing

The BraTS 2021 data were acquired using different clinical protocols, from different MRI scanners and multiple institutions, therefore, a pre-processing stage is essential. First, standard pre-processing routines have been applied by the BraTS challenge as stated in [2]. This includes conversion from DICOM into NIFTI file format, re-orientation to the same coordinate system, co-registration of the multiple MRI modalities, resampling to 1×1×1 mm isotropic resolution, and brain extraction and skull-stripping.

Following these pre-processing steps, we applied the image cropping stage where all brain pixels were cropped, and the resultant image was resized to a spatial resolution of 192×224×160. This method effectively results in a closer field of view (FOV) to the brain with fewer image voxels leading to a smaller resource consumption while training our deep learning models. Finally, z-score normalization was applied by subtracting the mean value and dividing by the standard deviation individually for each input MRI image.

### 2.3 Neural Network Architectures

We used two different CNN models, namely, DeepSeg [10] and nnU-Net [9] which follow the U-Net pattern [12, 13] and consist of encoder-decoder architecture interconnected by skip connections. The final results were obtained by using the Simultaneous Truth and Performance Level Estimation (STAPLE) [18] based on the expectation-maximization algorithm.

**DeepSeg.** Figure 2 shows a 3D enhanced version of our first model, DeepSeg, which is a modular framework for fully automatic brain tumor detection and segmentation. The proposed network differs from the original network in the following: First, the original DeepSeg network was proposed for 2D tumor segmentation using only FLAIR MRI data, however, we apply here 3D convolutions over all slices for more robust and accurate results. Second, we incorporate all the available MRI modalities (T1, T1Gd, T2, and T2-FLAIR) so that the GBM sub-regions could be detected in comparison with the whole tumor only in the original DeepSeg paper [10]. Third, we incorporate additional modifications such as region-based training, excessive data augmentation, a simple postprocessing technique, and a combination of cross-entropy (CE) and Dice similarity coefficient (DSC) loss functions.

Following the structure of U-Net, DeepSeg consists of two main parts: a feature extractor part and an image upscaling part. Downsampling is performed with $2 \times 2 \times 2$ max-pooling and upsampling is performed with $2 \times 2 \times 2$ up convolution. DeepSeg uses the recently proposed advances in CNNs including dropout, batch normalization (BN), and rectified linear unit (ReLU) [19, 20]. The feature extractor consists of five consecutive convolutional blocks, each containing two $3 \times 3 \times 3$ convolutional layers, followed by ReLU. In the image upscaling part, the resultant feature map of the feature extractor is upsampled using deconvolutional layers. The final output segmentation is attained using a $1 \times 1 \times 1$ convolutional layer with a softmax output.

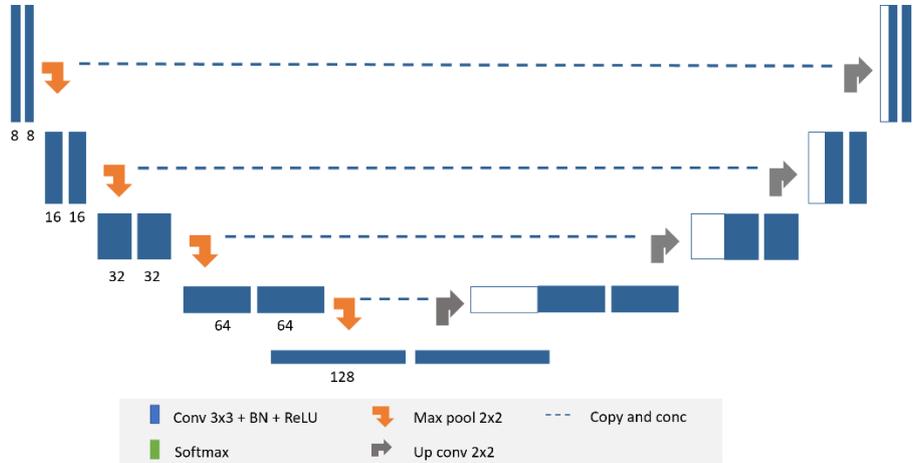

**Fig. 2.** DeepSeg network consists of convolution neural blocks (*blue boxes*), downsampling using maximum pooling (*orange arrows*), and upsampling using up convolution (*blue arrows*), and softmax output layer (*green block*). The input patch size was set to $128 \times 128 \times 128$.

**nnU-Net.** The baseline nnU-Net is outlined in Fig. 3, which is a self-adaptive deep learning-based framework for 3D semantic biomedical segmentation [9]. Unlike DeepSeg, nnU-net does not use any of the recently proposed architectural advances in deep learning and only depends on plain convolutions for feature extraction. nnU-Net used strided convolutions for downsampling and convolution transposed for upsampling. The initial filter size of convolutional kernels is set to 32 and doubled at the following layers with a maximum of 320 in the bottleneck layers.

By modifying the baseline nnU-Net and making BraTS-specific processing, nnU-Net won first place in the segmentation task of the BraTS challenge in 2020 [16]. The softmax output was replaced with a sigmoid layer to target the three evaluated tumor sub-regions: whole tumor (consisting of all 3 labels), tumor core (label 1 and label 4), and enhancing tumor (label 4). Further, the training loss was changed to a binary cross-entropy instead of categorical cross-entropy that optimized each of the sub-regions independently. Also, the batch size was increased to 5 as opposed to 2 in the baseline nnU-Net and more aggressive data augmentations were incorporated. Similar to DeepSeg, nnU-Net utilized BN instead of instance normalization. After all, the sample dice loss function was changed to batch dice by computing over all samples in the batch. In our experiments, we incorporated the top-performing nnU-Net configuration on the validation set of BraTS 2020.

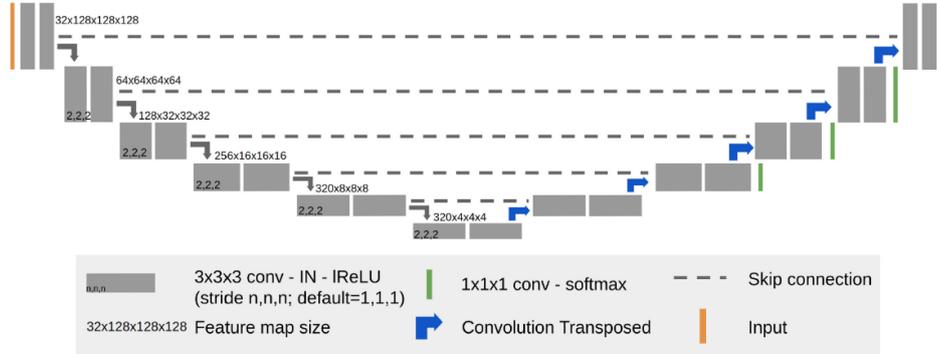

**Fig. 3.** nnU-Net network consists of strided convolution blocks (*grey boxes*), and upsampling as convolution transposed (*blue arrows*). The input patch size was set to 128 × 128 × 128 and the maximum filter size is 320 [16].

### 2.4 Post-processing

Determining the small blood vessels in the tumor core (necrosis or edema) is one of the most challenging segmentation tasks in the BraTS Challenge. In particular, this is clear in low-grade glioma (LGG) patients where they may not have enhancing tumors and, therefore, the BraTS challenge evaluates the segmentation as binary values of 0 or 1. Although if there are only small false positives in the predicted segmentation map of a patient with no enhancing tumor will result in a dice value of 0. To overcome this problem, all enhancing tumor output were re-labeled with necrotic (label 1) if the total predicted ET regions are less than a threshold. This threshold value was selected based on our analysis of the validation set results so that our model performs better. This strategy has a possible side effect of removing some correct predictions.

## 3 Experiments and Results

### 3.1 Cross-validation Training

We train each model as five-fold cross-validation on the 1251 training cases of BraTS 2021 for a maximum of 1000 epochs. Adam optimizer [21] has been applied with an initial learning rate of $1e^{-4}$ and a default value of $1e^{-7}$ for epsilon. Each configuration was trained on a single Nvidia GPU (RTX 2080 Ti or RTX 3060). The input to our networks is randomly sampled patches of 128 × 128 × 128 voxels with varying batch sizes from 2 to 5 and the post-processing threshold is set to 200 voxels. This tiling strategy allows the model to be trained on multi-modal high-resolution MRI images with low GPU memory requirements. The DeepSeg model was implemented using Tensorflow [22] while nnU-Net was implemented using PyTorch [23].

For training DeepSeg, the loss function is a combination of CE and DSC loss functions, which can be calculated as follows:

$$L_{DeepSeg} = DSC + CE = \frac{2*\sum yp + \varepsilon}{\sum y + \sum p + \varepsilon} - \sum y . \log(p) \quad (1)$$

where *p* denotes the network softmax predictions and $y \in \{0, 1\}$ representing the ground truth binary value for each class. Note that $\varepsilon$ is the smooth parameter to make the dice function differentiable.

To overcome the effect of class imbalance between tumor labels and the brain healthy tissue, we apply on-the-fly spatial data augmentations during training (random rotation between 0 and 30°, random 3D flipping, power-law gamma intensity transformation, or a combination of them).

### 3.2 Online Validation Dataset

The results of our models on the BraTS 2021 validation set are summarized in Table 1, where the five models for each cross-validation training configuration are averaged as an ensemble. Two evaluation metrics are used for the BraTS 2021 benchmark, computed by the online evaluation platform of Sage Bionetworks Synapse (Synapse), which are the DSC and the Hausdorff distance (95%) (HD95). We compute the averages of DSC scores and HD95 values across the three evaluated tumor sub-regions and then use them to rank our methods in the final column.

DeepSeg A refers to the baseline DeepSeg model, which has large input patches of the full pre-processed image, smaller batch size of 2. With DSC values of 81.64, 84.00, and 89.98 for the ET, TC, and WT regions, respectively, DeepSeg A model yields good results, especially when compared to the inter-rater agreement range for manual MRI segmentation of GDM [24, 25]. By using a region-based version of DeepSeg with an input patch size of 128 × 128 × 128 voxels, batch size of 5, applied post-processing stage, and on-the-fly data augmentation, the DeepSeg B model achieved better results of DSC values of 82.50, 84.73, and 90.05 for the ET, TC, and WT regions, respectively.

Additionally, we used two different configurations of the BraTS 2020 winning approach nnU-Net [16]. The first model, nnU-Net A, is a region-based version of the standard nnU-Net, large batch size of 5, more aggressive data augmentation as described in [16], trained using batch Dice loss, and including the postprocessing stage. nnU-Net B model is very similar to nnU-Net A model with applied brightness augmentation probability of 0.5 for each input modality compared with 0.3 for model A. nnU-Net models ranks second and third in our ranking (see Table 1) achieving an average DSC and HD95 results of 87.78, 87.87 and 9.6013, 10.1363 for each model, correspondingly.

For the RSNA-ASNR-MICCAI BraTS 2021 challenge, we selected the three top-performing models to build our final ensemble: DeepSeg B + nnU-Net A + nnU-Net B. Our final ensemble was implemented by first predicting the validation cases individually with each model configuration, followed by averaging the softmax outputs to obtain the final cross-validation predictions. After that, the STAPLE [18] was applied to aggregate the segmentation produced by each of the individual methods using the probabilistic estimate of the true segmentation. Our ensemble method is ranked among the top 10 teams for the BraTS 2021 segmentation challenge.

**Table 1.** Results of our five-fold cross-validation models on BraTS 2021 validation cases. All reported values were computed by the online evaluation platform Synapse. The average of DSC and HD95 scores are computed and used for ranking our methods.

| Model | DSC | | | | HD95 | | | | Rank |
|---|---|---|---|---|---|---|---|---|---|
| | ET | TC | WT | Avg | ET | TC | WT | Avg | |
| DeepSeg A | 81.64 | 84.00 | 89.98 | 85.21 | 19.77 | 10.25 | 5.11 | 11.71 | 5 |
| DeepSeg B* | 82.50 | 84.73 | 90.05 | 85.76 | 21.36 | 12.96 | 8.04 | 14.12 | 4 |
| nnU-Net A** | 84.02 | 87.18 | 92.13 | 87.78 | 16.03 | 8.95 | 3.82 | 9.60 | 2 |
| nnU-Net B*** | 83.72 | 87.84 | 92.05 | 87.87 | 17.73 | 8.81 | 3.87 | 10.14 | 3 |
| Ensemble (*, **, ***) | 84.10 | 87.33 | 92.00 | 87.81 | 16.02 | 8.91 | 3.81 | 9.58 | 1 |

### 3.3 Qualitative Output

Figure 3 shows the qualitative segmentation predictions on the BraTS 2021 validation dataset. These outcomes were generated by applying our ensemble model. The rows show the best, median, and worse segmentations based on their DSC scores, respectively. From this figure, it can be seen that our model achieves very good results with the overall high quality. Although the worst case, BraTS2021_Validation_01739, has a TC of zero, this finding was not quite surprising as illustrated in Section 2.4 as a side effect of applying our postprocessing strategy. Notably, the WT region was detected with a good quality (DSC of 95.72) which could be already valuable for clinical use.

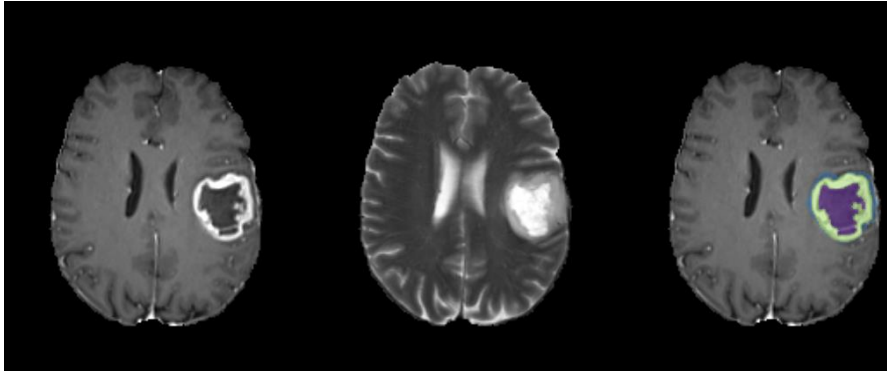

**Best:** BraTS2021_Validation_00153, EC (97.32), TC (98.77), WT (98.13)

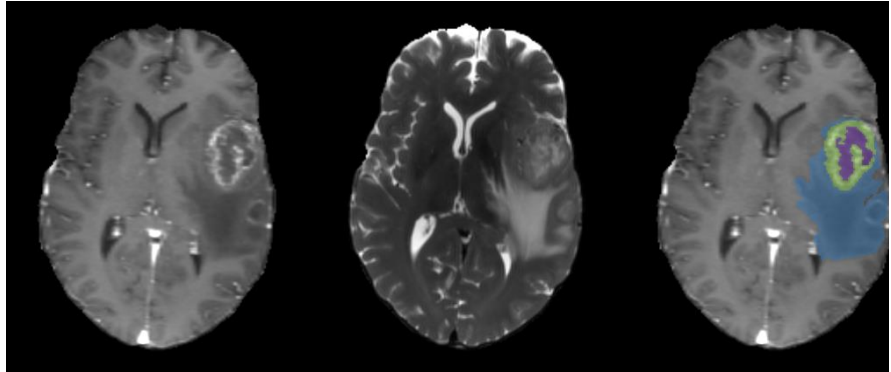

**Median:** BraTS2021_Validation_00001, EC (82.82), TC (91.04), WT (94.59)

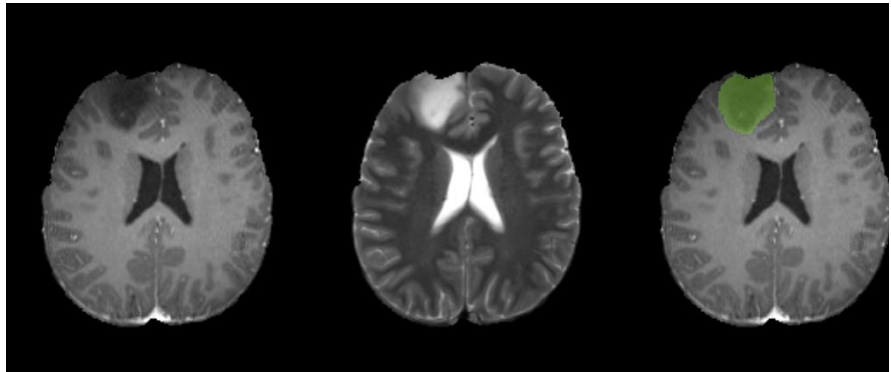

**Worst:** BraTS2021_Validation_01739, EC (0), TC (85.34), WT (95.72)

**Fig. 4.** Sample qualitative validation set results of our ensemble model. The best, median, and worse cases are shown in the rows. Columns display the T2, T1Gd, and the overlay of our predicted segmentation on the T1Gd image. Images were obtained by using the 3D Slicer software [17].

### 3.4 BraTS Test Dataset

Table 2 summarizes the final results of the ensemble method on the BraTS 2021 test dataset. Superior results were obtained for the DSC of ET, while all other obtained DSC results were broadly consistent with the validation dataset. In contrast, a substantial discrepancy between validation and test datasets for the HD95 is visible. Although our results were not state-of-the-art for the BraTS 2021 challenge, the proposed method showed better or equal segmentation performance to the manual inter-rater agreement for tumor segmentation [3]. The results confirm that our method can be used to guide clinical experts in the diagnosis of brain cancer, treatment planning, and follow-up procedures.

Table 2. Results of our final ensemble models on the BraTS 2021 test dataset. All reported values were provided by the challenge organizers.

|  | DSC | | | HD95 | | |
|---|---|---|---|---|---|---|
|  | ET | TC | WT | ET | TC | WT |
| Mean | 87.63 | 87.49 | 91.87 | 12.13 | 6.27 | 14.89 |
| StdDev | 18.22 | 24.31 | 10.97 | 59.61 | 27.79 | 63.32 |
| Median | 93.70 | 96.04 | 95.11 | 1.00 | 2.00 | 1.41 |
| 25quantile | 85.77 | 91.33 | 91.09 | 1.00 | 1.00 | 1.00 |
| 75quantile | 96.62 | 98.20 | 97.22 | 1.73 | 4.12 | 3.00 |

## 4     Conclusion

In this paper, we described our contribution to the segmentation task of the RSNA-ASNR-MICCAI BraTS 2021 challenge. We used an ensemble model of two encoder-decoder-based CNN networks namely, DeepSeg [10] and nnU-Net [16]. Table 1 and Table 2 list the results of our methods on the validation set and test set, respectively. Remarkably, our method achieved DSC of 92.00, 87.33, and 84.10 as well as HD95 of 3.81, 8.91, and 16.02 for, ET, TC, and WT regions on the validation dataset, respectively. For the testing dataset, our final ensemble yielded DSC of 87.63, 87.49, and 91.87 in addition to HD95 of 12.1343, 14.8915, and 6.2716 for ET, TC, and WT regions, correspondingly. These results ranked us among the top 10 methods for the BraTS 2021 segmentation challenge. Furthermore, qualitative evaluation supports the numerical evaluation showing a high-quality segmentation. Our clinical partner suggested that this approach can be applied for guiding brain tumor surgery.

**Acknowledgments.** The first author is supported by the German Academic Exchange Service (DAAD) [scholarship number 91705803].